\documentclass{article}

\usepackage{arxiv}

\usepackage[utf8]{inputenc} 
\usepackage[T1]{fontenc}    
\usepackage{hyperref}       
\usepackage{url}            
\usepackage{booktabs}       
\usepackage{amsfonts}       
\usepackage{nicefrac}       
\usepackage{microtype}      
\usepackage{lipsum}		
\usepackage{graphicx}
\usepackage{natbib}
\usepackage{doi}

\usepackage[linesnumbered,ruled,vlined]{algorithm2e}
\usepackage{graphicx}

\usepackage{times}
\usepackage{latexsym}

\title{Protagonists' Tagger in Literary Domain -- New Datasets and a Method for \textit{Person} Entity Linkage}

\author{ {\hspace{1mm}Weronika Łajewska} \\
	Faculty of Mathematics and Information Science\\
	Warsaw University of Technology\\
	Warsaw, Poland \\
	\texttt{weronikalajewska@gmail.com} \\
	\And
	\href{https://orcid.org/0000-0002-3407-7570}{\hspace{1mm}Anna Wróblewska} \\
	Faculty of Mathematics and Information Science\\
	Warsaw University of Technology\\
	Warsaw, Poland \\
	\texttt{anna.wroblewska1@pw.edu.pl} \\
}

\renewcommand{\shorttitle}{Protagonists' Tagger in Literary Domain}

\hypersetup{
pdftitle={A template for the arxiv style},
pdfsubject={q-bio.NC, q-bio.QM},
pdfauthor={David S.~Hippocampus, Elias D.~Striatum},
pdfkeywords={First keyword, Second keyword, More},
}

\begin{document}
\maketitle

\begin{abstract}
Semantic annotation of long texts, such as novels, remains an open challenge in Natural Language Processing (NLP). This research investigates the problem of detecting \textit{person} entities and assigning them unique identities, i.e., recognizing people (especially main characters) in novels. We prepared a method for \textit{person} entity linkage (named entity recognition and disambiguation) and new testing datasets. The datasets comprise 1,300 sentences from 13 classic novels of different genres that a novel reader had manually annotated. Our process of identifying literary characters in a text, implemented in \textit{protagonistTagger}, comprises two stages: (1) named entity recognition (NER) of persons, (2) named entity disambiguation (NED) -- matching each recognized person with the literary character's full name, based on \textit{approximate text matching}. The \textit{protagonistTagger} achieves both precision and recall of above 83\% on the prepared testing sets. Finally, we gathered a corpus of 13 full-text novels tagged with \textit{protagonistTagger} that comprises more than 35,000 mentions of literary characters. 
\end{abstract}

\section{Introduction}
Novels are a fascinating field of study, not only for philologists or literary scholars but also for scientists involved in NLP. They are an excellent repository of knowledge about the language, people and their relations, historical events, places, expected behaviours, etc. Analysis of novels can be applied to detecting the relations between protagonists, creating summaries, location detection, creating timelines of events, and many more. \par

The first type of annotations that is crucial when discussing novels is related to protagonists. Marking all appearances of the novel's main characters is vital in the corpus to make a more advanced analysis. Due to many ambiguities appearing in novels, their complex structure, and often a broad spectrum of protagonists, this task is challenging. Furthermore, there are no annotated sets that can be used for testing or training models for this specific task in the literary domain. People's recognition or disambiguation in novels is similar to non-literary texts, such as newspaper articles or blogs.  However, in some aspects, this task is more complex in the literary domain. To mention a few of them: (1) dialogues in which the form of the name depends on the person addressing given protagonist (for example, using diminutive), (2) a vast amount of literary characters, very often sharing the same surname and (3) surprising names of protagonists (for example, \emph{the Creature} in \emph{Frankenstein} or \emph{Black Dog} in \emph{Treasure Island}). \par

In most use cases, it may not be enough to annotate each literary character with a general tag \emph{person}. To be able to analyze the novel on deeper levels, we need contradistinction between protagonists. The most desired way is to have a unique identity for each protagonist and assign it to this literary character's appearance in a text. Ideally, each instance should be associated/linked with a tag containing the full name of a protagonist along with her/his personal title to identify and differentiate literary characters precisely~\cite{entity_linkage}. Extracting parts of the novels associated with specific characters makes an in-depth analysis of long texts, such as sentiment-based analysis of individual protagonists, possible (see~Section~\ref{related_works:applications}). \par 

An example of such annotated text is given in~Table~\ref{tab:exemplary_output}. Our prepared tool -- called \emph{protagonistTagger} -- works automatically with a list of protagonists' names and a text of the novel given as input. 
\begin{table}[!htb]
\centering
\footnotesize
\begin{tabular}{ p{15cm} }
\toprule
    \emph{"Her disappointment in \textbf{Charlotte\textsubscript{<<Charlotte Lucas>>}} made her turn with fonder regard to her sister, of whose rectitude and delicacy she was sure her opinion could never be shaken, and for whose happiness she grew daily more anxious, as \textbf{Bingley\textsubscript{<<Charles Bingley>>}} had now been gone a week and nothing more was heard of his return. \textbf{Jane\textsubscript{<<Jane Bennet>>}} had sent \textbf{Caroline\textsubscript{<<Caroline Bingley>>}} an early answer to her letter and was counting the days till she might reasonably hope to hear again. The promised letter of thanks from \textbf{Mr. Collins\textsubscript{<<Mr. William Collins>>}} arrived on Tuesday, addressed to their father, and written with all the solemnity of gratitude which a twelve month's abode in the family might have prompted." } \\
\bottomrule
\end{tabular}
 \caption{An exemplary text extracted from novel \emph{Pride and Prejudice} by Jane Austen. Correct tags are written in a subscript of each recognized named entity of category \emph{person}.}
\label{tab:exemplary_output}
\end{table}

\noindent Our main contributions in this research include:
\begin{itemize}
    \item verification of the standard NER model performance in the literary domain and describing a method for preparing training sets for model fine-tuning (Section~\ref{sec:ner}),
    \item the \emph{protagonistTagger} tool for recognition and disambiguation of \textit{person} entities (entity linkage) in literary texts (Section~\ref{sec:method}),
    \item entity linkage benchmark testing datasets for entities of category \textit{person} manually annotated also with full names of literary characters (1,300 sentences each) (Section~\ref{sec:testing_sets}),
    \item a vast corpus of novels annotated with \textit{protagonistTagger} (13 novels, more than 50,000 sentences and more than 35,000 mentions of literary characters) (Section~\ref{sec:corpus}).
\end{itemize}
The tool's good performance verified on two different testing sets proves that \textit{protagonistTagger} can be used successfully to recognize and identify people in complex texts. The tool and the corpus can help detect and analyze the relationships between literary characters and create a social semantic network of them. Furthermore, the fine-tuned NER model and the whole tool can be applied to texts from a non-literary domain, containing named entities of category \emph{person}, as long as a list of full names to be linked with them is available. \par

In this paper, we provide an overview of the related works (Section~\ref{sec:related_work}), followed by the general description of the approach and the created tool -- \textit{protagonistTagger} (Section~\ref{sec:method}). All prepared testing sets are presented in~Section~\ref{sec:testing_sets}. Section~\ref{sec:ner} describes NER models used, whereas Section~\ref{sec:matching_alg} presents the name \textit{matching algorithm} of literary characters. The performance of the approach presented in this paper is described in Section~\ref{sec:experiments_evaluation}. Section~\ref{sec:corpus} introduces a corpus of full-text novels with mentions of literary characters. The paper completes with conclusions and future works (Section~\ref{sec:conclusions}). The detailed analysis of datasets, parameters for models reproducibility and other detailed analysis and statistics regarding this paper are given in the appendixes.\footnote{\footnotesize{The source code of \textit{protagonistTagger}, the corpus of 13 annotated full-text novels and manually annotated datasets are available, along with documentation and manual, on~\url{https://doi.org/10.5281/zenodo.4699418}.}}

\section{Related Works}\label{sec:related_work}
Literary text analysis has been performed for centuries by humanists and linguists in a manual, arduous way. Now, this field is offered a brand new perspective thanks to digital literary studies. Computational linguistics makes tasks such as in-depth statistical analysis of literary texts much quicker (less labour-intensive) and more precise. \par 

\subsection{Named Entity Recognition and Disambiguation}
The general conclusion can be drawn that standard NER and NED models can be used successfully on dissected datasets. However, they may have lower performance on novels containing veritable sentences~\cite{ner_glass_ceiling}. Further, we show these challenges in~Sections~\ref{sec:ner} and~\ref{sec:matching_alg}. 

NER task can be approached in several different ways~\cite{survey_2018_yadav}. The most recent approaches are designed using mainly neural networks. They do not require domain-specific resources or feature engineering, thanks to word embeddings used as feature vectors~\cite{embeddings_ner}. Neural network-based NER systems can be classified into several groups depending upon their representation of the words in a sentence. Such representations can be based on words~\cite{nlp_from_scratch_2011}, characters~\cite{ner_char,lstm_2016,nn_2016}, sub-word units different than characters~\cite{affix_2018} or any combination of these. This last approach, originating from feature engineering, where affixes play a significant role, offers an exhaustive way of creating word representations utilizing the semantics of morphemes. \par 

An up-to-date list of entity linkage techniques and datasets are given in \cite{nlp_progress, ned}. However, these resources are concentrated on texts such as news and most models are not prepared to face particular challenges appearing in literary texts.

\subsection{Recognition and Disambiguation of the Literary Characters}
The most popular library of texts of various kinds is \emph{Project Gutenberg} ~\cite{project_gutenberg}. It is accompanied by GutenTag~\cite{gutentag,gutentag_2} -- a software tool offering NLP techniques for the analysis of literary texts. It contains an automatic corpus reader, subcorpus filters and a tagging functionality based on different tags specified by a user, performing statistical analysis of the selected subcorpus. Even though the NER is used to identify the main literary characters, the information gained is used only for statistical measures. There is no advanced mechanism of recognition and disambiguation of literary characters. \par

One of the possible approaches to detecting and matching characters in literary texts, based on characters clustering~\cite{bayesian_character_types,extracting_social_networks}, assumes that the noun phrases recognized in the text by the NER model can be clustered into groups referring to the same person. In the case of both papers, the presented results are more qualitative than numeric and they are based on verifying preregistered hypotheses. Another method~\cite{detecting_characters} is based on representing characters using a graph, where each node corresponds to a name found using NER and edges connect nodes referring to the same character. Furthermore, this method attempts to identify entities that the NER has not recognized by uncovering prototypical characters' behaviours. Accuracy of character detection using this method varies between 45\% and even 75\%, depending on the novel and the challenges appearing in it. \par

Even though numerous papers devoted to recognizing literary characters are quoted here, in most cases there are no available datasets to compare the results or models' parameters to recreate the experiments~\cite{extracting_social_networks}. Moreover, some of the proposed methods are applicable only to some narrow set of texts and require a manual contribution. A good quality, general-purpose method for long, complex texts such as novels and benchmark datasets would be of great applicability.

\subsection{Further Applications Based on Recognized Literary Characters}\label{related_works:applications}
Constructing the representation and interpretation of narratives, extracting social networks from novels, modelling social conversations that occur between characters in the form of a network~\cite{extracting_social_networks} are promising directions when a corpus with recognized protagonists is available. Besides, literary characters and their relationships evolve with the progress of the novel. Modelling dynamic relationships between pairs of characters by detecting relationship sequences in data is much more adequate in the case of long texts~\cite{evolving_relationships,unsupervised_evolving_relationships}. 

Another aspect of the interpretation of narratives is sentiment analysis~\cite{sentiment_analysis_survey}. It includes, among others, a classification of literary texts by literary genre based on emotions they convey~\cite{emotional_arc_of_stories,happy_endings} and emotion-based character analysis~\cite{IR_folktales,personality_profiling}. \par 

\section{\emph{ProtagonistTagger} Workflow}
\label{sec:method}
\begin{figure}
    \centering
    \includegraphics[width=0.5\textwidth]{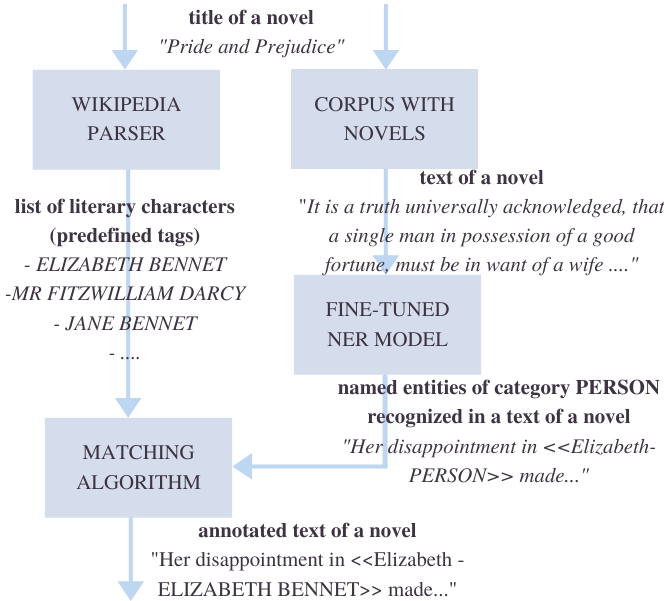}
    \caption{Simplified process employed in our tool \emph{protagonistTagger} with examples.}
    \label{fig:workflow_part_1}
\end{figure}

The process of creating the corpus of annotated novels employed in the \emph{protagonistTagger} tool comprises several stages (see~Figure~\ref{fig:workflow_part_1}):
\begin{enumerate}
    \item Gathering an initial corpus with plain novels' texts without annotations.
    \item Creating a list of full names of all protagonists for each novel in the initial corpus. These names are the predefined tags that will be used in further steps for annotations. This step uses \emph{Wikipedia} parser, which goes through an article about a given novel looking for a section devoted to its literary characters. \label{schema:preparing_labels}
    \item Recognizing named entities of category \emph{person} in the novels' texts in the initial corpus. Training the NER model from scratch for this specific problem is not reasonable due to the amount of time and computing power required. Instead, we used a pretrained NER model and fine-tuned it using a sample of manually annotated texts. The NER evaluation is done on a testing set extracted from the novels. \label{scheme:NER_fine_tune}
    \item Each named entity of category \emph{person}, recognized by the NER model in the previous step, is a potential candidate to be annotated with one of our tags predefined in step~\ref{schema:preparing_labels}. At this point, the \textit{matching algorithm} is used to choose from the predefined tags the one that matches most accurately the recognized named entity. \label{schema:autotager}
    \item The annotations done by the \emph{matching algorithm} are evaluated according to their accuracy and correctness.
\end{enumerate} \label{schema:steps}
The \emph{protagonistTagger} (fine-tuned NER model + \textit{matching algorithm}) is used to annotate more novels in order to create the corpus of annotated texts. The two most important parts of the above procedure are fine-tuning NER from step~\ref{scheme:NER_fine_tune} and the \emph{matching algorithm} from step~\ref{schema:autotager}. They are described in detail in the two following chapters.

\begin{table}
\footnotesize
    \centering
 \setlength{\arrayrulewidth}{0.1mm}
\setlength{\tabcolsep}{3pt}
\renewcommand{\arraystretch}{1}
    \begin{tabular}{  p{2.5cm}  p{0.75cm}  p{0.75cm}  p{0.75cm}  p{2cm}  }
    \textbf{Testing set} & \textbf{\# spn} & \textbf{\# n.} & \textbf{size} & \textbf{Tags} \\ [0.5ex]
    \hline
     \textit{Test\_large\_person} & 100 & 10 & 1,000 & \emph{person} \\
     \textit{Test\_large\_names} & 100 & 10 & 1,000 & full names \\
     \textit{Test\_small\_person} & 100 & 3 & 300 & \emph{person} \\
     \textit{Test\_small\_names} & 100 & 3 & 300 & full names \\
    \end{tabular}
        \caption{Summary of all testing sets used. For every set, the gold standard annotations are indicated manually, either with general tag \emph{person}, or with full names of literary characters. Note: \#spn -- number of sentences per novel, \# n. -- the number of novels, size -- the number of all sentences in a given dataset.  }
    \label{tab:testing_sets}
\end{table}

\section{Our Testing Sets}\label{sec:testing_sets}
Two testing sets are prepared for verifying the performance of the NER model. The novels included in \textit{Test\_large\_person} are also used in the training data for the NER model; however, the training sets and the testing sets are disjoint. Additionally, the same novels were used to verify which named entities are not recognized and classified correctly by the standard NER model. Taking this into consideration, the results on this dataset may not be entirely trustworthy. Therefore, additional independent testing set containing sentences from totally new novels was created -- \textit{Test\_small\_person}. The results of the NER model on this testing set are undoubtedly authoritative. The sentences from both testing sets contain various named entities of the category \textit{person}. The gold standard for both datasets was annotated manually, by a passionate novel reader, with a tag \textit{person}. \par 

The overall performance of \emph{protagonistTagger} is evaluated on two testing sets: \emph{Test\_large\_names} and \emph{Test\_small\_names}. Both testing sets include the same sentences as the corresponding sets used for testing the NER model. The only difference is that this time the sentences are manually annotated with full names of literary characters while creating the gold standard. Table~\ref{tab:testing_sets} sums up all the testing sets. \par

All testing sets contain sentences chosen randomly from 13 novels differing in style and genre. The testing sets used for \textit{protagonistTagger} contain all together 1,300 sentences (100 sentences from each novel). Each sentence in the testing set contains at least one named entity of category \emph{person} recognized by the NER model. We need to bear in mind that not all literary characters appearing in a novel are given on Wikipedia. Therefore, not all literary characters appearing in the novel are included in the predefined tags (the lists of protagonists extracted from Wikipedia). Nevertheless, sentences containing some names of minor literary characters not included in the tag lists are also included in the testing set. They should be given only a general tag \emph{person}. It helps to verify if some additional, undesired annotations are not put in by our \textit{protagonistTagger}.

\section{NER Models}\label{sec:ner}
The novel is a particular type of text in terms of writing style, the links between sentences, the plot's complexity, the number of characters, etc. Therefore, we needed to verify the standard NER model's performance on exemplary sentences extracted from novels. It is mainly because all the standard NER models are pretrained on web data such as blogs, news, and comments \cite{spacy_eval_2,spacy_eval}. We tested several NER models on literary domain and based on preliminary results, we decided to use a pretrained language model offered by SpaCy.\footnote{\footnotesize{\url{https://spacy.io}}} 

\subsection{Standard NER Model Performance}
The overall recall metric on the whole testing set \emph{Test\_large\_person} for the pretrained NER model was 80\% (see~Table~\ref{tab:metrics_fine_tuned_ner}). In some of the tested novels, the most alarming thing is that the NER model does not recognize the main protagonist's name as an entity of category \emph{person}. It is the case in \emph{The Picture of Dorian Gray}, where the entity \emph{Dorian} is recognized as \emph{norp} -- nationalities or religious or political groups, and in the novel \emph{Emma}, where the entity \emph{Emma} is given a label \emph{org} that should be assigned to companies, agencies, or institutions. 

However, we need to bear in mind that NER is the first part of the protagonists' annotation process and there is still a potential error of the further steps, i.e., \textit{matching algorithm}. The overall performance of the \emph{protagonistTagger} on a novel drops drastically, even with an excellent process of NED, when the main protagonist's name is not detected correctly. Therefore, it is crucial to fine-tune the NER model so that it can handle the recognition of these entities.

\subsection{NER Model Fine-Tuning}\label{subsec:ner-fine-tuning}
NER model aims at finding all named entities of category \emph{person} which may be matched in the further step with the proper label (i.e., a full name of a novel's protagonist). Therefore, we want the NER model to find as many \emph{person} entities as possible to have the highest possible number of candidate entities for the matching phase. It means that the recall metric is high enough. The NER model's precision at this step is not crucial because the named entities of the other categories, for example, \emph{location}, identified as \emph{person}, are not matched with any protagonist's label in the matching phase. The \emph{matching algorithm} ignores (filters out) named entities that do not resemble any of the predefined tags. 

\subsubsection{NER Model Fine-Tuning Procedure} 
Fine-tuning the NER model is used to improve its performance on novels~\cite{transfer_learning}. As a base for fine-tuning, we use an existing, pretrained language model.

This procedure comprises the following steps:
\begin{enumerate}
        \item \label{schema:first_step_ner} NER model is applied to a testing set extracted from the novels. The resulted annotations are compared with the testing set annotated manually with the general tag \emph{person}. 
        \item If the results of NER are not satisfying (many entities that are of category \emph{person} are not recognized, or they are assigned a different category), then the NER model needs to be fine-tuned. Fine-tuning NER requires creating a training set with texts specific to our problem (i.e., novels). 
\end{enumerate}

\subsubsection{Training Sets for NER Fine-Tuning} \label{NER_training_sets}
We considered two approaches to creating a training set for a NER model~\cite{transfer_learning_training_sets}. The final training set for NER model fine-tuning is the concatenation of \textit{Training\_set\_1} and \textit{Training\_set\_2}. The performance of the fine-tuned NER model is discussed in~Section~\ref{performance_ner}. \par
\textit{Training\_set\_1} contains 485 sentences with not recognized named entities of category \emph{person}. This approach assumes using the named entities of type \emph{person} that were not recognized or assigned a proper category during the analysis of the problem (these named entities were found manually). The novels used for creating the training set are scanned in search of sentences containing these entities. Then the chosen sentences are annotated in a semi-automatic way with a general tag \textit{person} creating a training set for fine-tuning the standard NER model. \par

\begin{table}
    \footnotesize
    \centering
    \begin{tabular}{ p{15cm}  }
    \toprule
            \emph{"\textbf{Jane}'s delicate sense of honour would not allow her to speak to \textbf{Elizabeth} privately of what \textbf{Lydia} had let fall; \textbf{Elizabeth} was glad of it; till it appeared whether her inquiries would receive any satisfaction, she had rather be without a confidante."} \\ 
    \vspace{0.1cm}
        \emph{"\textbf{Deborah}'s delicate sense of honour would not allow her to speak to \textbf{Harvey} privately of what \textbf{Lydia} had let fall; \textbf{Harvey} was glad of it; till it appeared whether her inquiries would receive any satisfaction, she had rather be without a confidante."} \\ 
    \bottomrule
    \end{tabular}
    \caption{An example of injecting common English names in the sentence extracted from \emph{Pride and Prejudice} by Jane Austen. In this case \emph{Jane} is replaced by \emph{Deborah} and \emph{Elizabeth} is replaced by \emph{Harvey}.}
    \label{tab:name_replacement}
\end{table}

\textit{Training\_set\_2} contains 1,600 sentences from novels with injected common English names. Many common English names, such as \emph{Emma, Charlotte, Arthur} or \emph{Grace}, are not recognized at all by a standard NER model, or they are classified as entities of a type different than \emph{person}. For the NER model to recognize common names and thus improve its performance, the training set needs to contain sentences typical for novels regarding style, vocabulary, and syntax. These sentences should additionally contain as many common English names as possible. The easiest way to create such a set of correct sentences is to extract from novels sentences containing the main protagonists' names. Then each such name can be replaced by some other common English name to enrich our training set, employing the approach described in~\cite{ner_glass_ceiling}. An example of such replacement is given in~Table~\ref{tab:name_replacement}. The considered list of most common English names contains 300 female and 300 male names in their basic forms.

\section{Name \textit{Matching Algorithm}}
\label{sec:matching_alg}

The \textit{matching algorithm} matches all protagonists in a given text with a proper tag (i.e., a proper name), having been given the list of protagonists' proper names predefined for each novel. The algorithm evaluates the match between the recognized named entity and each full name from this predefined list. The tag with the highest resemblance is chosen as an answer. The method is mainly based on \emph{approximate string matching}. The algorithm attempts to solve several particular problems encountered during the problem analysis, such as diminutives or surnames preceded with personal titles. Encountered problems along with the proposed solutions are presented in~Sections~\ref{partial_string_matching},~\ref{problems_dimunitives}~and~\ref{problems_personal_title}. The matching algorithm, of course, cannot handle all of the possible cases. It would be highly ineffective due to complexity issues. However, the problems appearing most frequently in the analyzed novels are considered. 

We assumed in our approach that we possess a list of protagonists' full names for the processed novel (either thanks to Wikipedia parser or by creating it manually and proving as an input for the tool in case of novels that do not have Wikipedia entries). The task is to recognize the named entities in the text and match them appropriately with labels from this list (named entities disambiguation). It is crucial to point out that the named entities in the novel's text rarely take the same form as in the list. Sometimes only the first name is used; in other cases, the surname preceded with a personal title appears. In extreme cases, a diminutive or a nickname may be used (for example, \emph{Lizzy} instead of \emph{Elizabeth} or \emph{Nelly} instead of \emph{Ellen}). \par

Considering the above, it needs to be verified how similar a named entity is to each label from the list. Only then will it be possible to assign a proper, most similar label to the entity. A technique called \textit{approximate text matching} can be used to calculate this similarity. The problem of \emph{approximate text matching} can be formally stated as follows: given a long text \(T_{1, ..., n}\) of length \emph{n} and a comparatively short pattern \(P_{1, ..., m}\) of length \emph{m}, both sequences over an alphabet \(\sum\) of size \(\rho\), find the text positions that match the pattern with at most \emph{k} "errors"~\cite{navarro2001indexing}.

\subsection{Regular and Partial String Matching}\label{partial_string_matching}
String similarity, based on \textit{approximate text matching}~\cite{navarro2001guided}, can be computed in multiple different ways. The general method uses Levenshtein distance to calculate differences between two sequences of characters. Formally speaking, the distance/error \(d(x, y)\) between two strings \emph{x} and \emph{y} is the minimum number of single-character operations (such as insertion, deletion, and substitution) needed to convert one into the other. \par

In the context of our problem, the basic measurement of Levenshtein distance between a named entity found in a text and a character name from a list may not solve the problem. Using this method for entity \emph{Elizabeth} and full name \emph{Elizabeth Bennet} gives the similarity of only 72\%. A modification of the basic method, which calculates the so-called \emph{partial string similarity}, gives the similarity of 100\%, which is precisely what we expected (see~Table~\ref{tab:ASM}). This modification uses a heuristic called \emph{best partial} which, given one sequence of length \emph{n} and a noticeably shorter string of length \emph{m}, calculates the score of the best matching substring of length \emph{m} of the sequence.

\begin{table}
\centering
\footnotesize
 \setlength{\arrayrulewidth}{0.1mm}
\setlength{\tabcolsep}{3pt}
\renewcommand{\arraystretch}{1}
    \begin{tabular}{  p{1.7cm}  p{2.8cm}  p{1.1cm}  p{1.1cm} }
    \textbf{Named entity} & \textbf{Literary character's full name} & \textbf{Regular str. sim.} & \textbf{Partial str. sim.}\\ [0.5ex]
    \hline
     Elizabeth & Elizabeth Bennet & 72\% & 100\% \\
     Lizzy & Elizabeth Bennet & 19\% & 40\% \\
     Lizzy & Mr Fitzgerald Darcy & 24\% & 40\% \\
    \end{tabular}
        \caption{Examples of calculated string similarities (str. sim.) for some of the named entities recognized in the novel \emph{Pride and Prejudice}.}
    \label{tab:ASM}
\end{table}

\subsection{Diminutives of Literary Characters}\label{problems_dimunitives}
The most difficult cases in the matching process are diminutives and nicknames. The problem accompanies the character \emph{Elizabeth Bennet}, who is sometimes called \emph{Lizzy} by her family. We do not have any information about the possible forms of the name in our list of labels, which contains only the base forms. \emph{Lizzy} is recognized as a named entity of category \emph{person}, but it is not similar enough to any of the protagonists from the list. The partial string similarity between \emph{Lizzy} and \emph{Elizabeth Bennet} equals 40\% and is the same as between \emph{Lizzy} and \emph{Mr Fitzwilliam Darcy} (see~Table~\ref{tab:ASM}). Therefore, the straightforward approximate string matching technique is not enough in this case. Instead, the complete list of diminutives containing more than 3300 different forms of names is used. We investigate it only when the recognized named entity is not similar enough to any of the protagonists listed in a list of labels for a given novel. \par

\subsection{Surnames Preceded with Personal Title}\label{problems_personal_title}
Another case that needs special consideration is a named entity consisting only of a surname. 
For example, the named entity \emph{Bennet} is not the name of any specific character, but instead the whole family's name. To distinguish between \emph{Bennet} meaning the whole family and \emph{Bennet} being the surname of a single character, we can analyze the word preceding it. \emph{Bennet} preceded with a personal title such as Mr., Mrs., Ms. or Miss, should be identified as a single person, whose surname is \emph{Bennet}. Additionally, each prefix stores valuable information about the gender of the literary character that should be assigned to the analysed entity. In all other cases, \emph{Bennet} is treated as the whole family and not a single person identified in a text. For example, the entity \emph{Bennet} appears 323 times in the novel (which has 121,533 words), out of which 314 cases can be analyzed more precisely thanks to the preceding personal title. \\

\subsection{\textit{Matching Algorithm} Outline} \label{matching_algorithm_outline}
The general idea behind the \textit{matching algorithm} is finding the best match for a recognized named entity of category \emph{person} in the predefined list of considered protagonists appearing in the novel. First, it collects potential candidates that may correspond to the given named entity from the protagonists' list. Then the algorithm, based on the \emph{approximate string matching} method, chooses the best match from this list of potential candidates. 

\section{Experiments and Evaluation}\label{sec:experiments_evaluation}

\subsection{Fine-tuned NER Model Performance}\label{performance_ner}
We have tested the standard NER model and the standard NER model fine-tuned with the prepared training set (see~Section~\ref{NER_training_sets}). Performance of these NER models on the \emph{Test\_large\_person} is presented in~Table~\ref{tab:metrics_fine_tuned_ner}. At this point, we were interested mostly in the recall (see~Section~\ref{subsec:ner-fine-tuning}). We fine-tuned NER in order to be able to detect most of the named entities of category \emph{person}. The obvious conclusion from the results on this testing set is that the recall for the standard NER model, in almost all cases, is significantly lower than for the fine-tuned model. It confirms our hypothesis that the standard NER models are not prepared for novels. \par 

The performance of all NER models on the \emph{Test\_small\_person} makes us draw similar conclusions. The fine-tuned model has a higher recall on the whole testing set in general and in the case of almost every novel analyzed individually.
 What is important, its performance on a new testing set is also very high. Furthermore, the results on this smaller set are more reliable and independent, due to the fact that the larger set is extracted from novels used in problem analysis and NER model fine-tuning (even though training set and testing sets are disjoint). \par

\begin{table}[!h]
 \footnotesize
 \centering
 \setlength{\arrayrulewidth}{0.1mm}
\setlength{\tabcolsep}{3pt}
\renewcommand{\arraystretch}{1}
\begin{tabular}{ p{1.6cm}  p{1.3cm}  p{1.1cm}  p{1.5cm}  p{1.2cm}}
\textbf{NER model} & \textbf{Precision} & \textbf{Recall} & \textbf{F-measure} & \textbf{Support}\\
\hline
\multicolumn{5}{c}{\emph{Test\_large\_person}} \\
\hline
    standard &  0.84 &  0.8  &  0.82 &  1,021 \\
    fine-tuned &  0.77 &  0.99 &  0.87 &  1,021 \\
\hline
\multicolumn{5}{c}{\emph{Test\_small\_person}} \\
\hline
    standard  & 0.78 &  0.79 &  0.78 &  273 \\
    fine-tuned &  0.69 & 0.95 &  0.8  & 273 \\
\end{tabular}
\caption{Metrics computed for the standard, pretrained NER model and the fine-tuned NER model for annotations with general label \emph{person}. The \emph{support} is the number of occurrences (mentions) of class \emph{person}.}
\label{tab:metrics_fine_tuned_ner}
\end{table}

\subsection{\textit{ProtagonistTagger} Performance}\label{performance_pt}
All the named entities of category \emph{person} are included in the performance, not only the main characters' names. Therefore, the performance addresses recognizing and annotating the most important protagonists in the novel appearing in the predefined tags and the minor, tangential ones. Generally, the tool achieves high results on both testing sets -- precision and recall above 83\% (see~Table~\ref{tab:full_tags_large_set_overall_metrics}). The tool's performance on \emph{Test\_small\_names} shows that it can be successfully used for new novels to create a larger corpus of annotated texts. \par

\begin{table}
 \footnotesize
 \centering
 \setlength{\arrayrulewidth}{0.1mm}
\setlength{\tabcolsep}{3pt}
\renewcommand{\arraystretch}{1}
\begin{tabular}{ p{2.4cm} p{1.5cm} p{1.5cm} p{1.5cm} }
 \textbf{Testing set} & \textbf{Precision} & \textbf{Recall} & \textbf{F-measure} \\
\hline
\emph{Test\_large\_names}   &    0.88 &     0.87 &       0.87 \\
\emph{Test\_small\_names}   &    0.83 &     0.83 &       0.83 \\
\end{tabular}
\caption{Performance of the \emph{protagonistTagger}.}
\label{tab:full_tags_large_set_overall_metrics}
\end{table}

It should be emphasized that the \textit{novel} is a term defining a wide range of various texts. The best proof of novels' diversity is the tool's performance, whose precision varies from 79\% to even 96\% for different novels. Even though the performance is tested on various texts, the precision of the annotations remains high, proving the applicability of the proposed method in the literary domain. \par

\subsection{Experiment Limitations}
Considering only the novels included in the testing sets, it can be concluded that there are numerous factors negatively influencing the performance of the \emph{protagonistTagger}. It depends on the number of literary characters appearing in the novel, its complexity, literary genre, uniqueness of predefined tags, and probably many more. 
The main challenge is to assign tags to named entities appearing in a form common for the novel (for example, only surnames preceded with the personal title or many diminutives). What is crucial, this last dependency is novel-specific and author-specific. 

\section{Corpus of Annotated Novels}\label{sec:corpus}
The \textit{protagonistTagger} was employed to create a corpus of annotated novels. Named entities of category \textit{person} are annotated in the texts with the full names of the corresponding literary characters. The corpus contains 13 novels (altogether more than 50,000 sentences) and more than 35,000 annotations of literary characters. The annotations done on this corpus are guaranteed to be of good quality (precision and recall on average above 83\%) because the annotation tool was tested on the independent testing sets extracted from the novels included in this corpus.  \par

\section{Conclusions and Further Work}\label{sec:conclusions}
In this paper, we presented the tool \emph{protagonistTagger} and the corpus of annotated novels where each literary character is tagged with her/his full name. The \textit{protagonistTagger} achieved both the precision and the recall of above 83\% on the testing sets containing texts from thirteen novels. This research shows the relatively low performance of the standard NER models on novels for recognizing entities of category \emph{person}. It is caused by the fact that novels differ significantly from texts on which standard NER models are trained. Our contribution was to describe the method for preparing training sets and fine-tuning the NER model. It resulted in a very high recall (above 95\%) in recognizing named entities of category \emph{person} in novels. \par

Further contributions concern a method for matching appearances of protagonists in texts of the novels to their full names, i.e., linking the recognized \textit{person} named entity with the literary character's identity. The analysis of the problem unveiled this process's complexity. Thus, even though the tool \emph{protagonistTagger} partially builds upon existing approaches and resources, it points out the challenges of the task in the literary domain and adapts the existing materials for it. Furthermore, we introduce new benchmark sets for NER and NED in the literary domain, i.e., datasets for tuning testing NER models and testing NED, and the corpus annotated automatically with \textit{protagonistTagger}.

The variety of possibilities presented in~Section~\ref{sec:related_work} makes further analysis of novels based on the created corpus very tempting. The created corpus and the tool for annotating new novels seem to be a good starting point in many NLP tasks. We also performed initial tests to apply our results, i.e., analyzing sentiment and relationships in novels using our corpus. The experiments' main goal was to determine each character's sentiment in terms of positive, neutral, and negative and the degree of affiliation to a specific class. In addition, we examined how the character's sentiment changes over the time of the novel. Furthermore, we attempted to discover how relationships between the two characters develop over time. The performed analysis was more manageable and precise, thanks to the available annotations of literary characters than the standard tool used so far.

The \emph{protagonistTagger} was created initially for a literary domain, especially novels. Nevertheless, it has the potential to be easily applied to other kinds of texts in which we want to annotate \emph{person} named entities. The only condition is the access to the predefined tags defining the full names to be matched with named entities detected in a text. Texts extracted from social media very often feature many names in different forms. Therefore, recognizing and annotating them can be very beneficial from the point of view of investigating human opinions and analyzing the sentiments.

\bibliography{custom}
\bibliographystyle{acl_natbib}

\appendix
\section*{Appendices}
\addcontentsline{toc}{section}{Appendices}
\renewcommand{\thesubsection}{\Alph{subsection}}

\section{\textit{Matching Algorithm} Pseudocode} \label{appendix:matching_alg}
The algorithm takes as inputs:
\begin{itemize}
    \item \textbf{named\_entity} -- a named entity of category \emph{person} found by NER model,
    \item \textbf{protagonists} -- a predefined list of all considered literary characters/protagonists,
    \item \textbf{prefix} -- a prefix that is a token appearing before the recognized named entity; it can be a personal title, the article \emph{the}, or an empty string,
    \item \textbf{partial\_similarity\_precision} -- value indicating how similar two strings need to be in order to be considered as potential matches; it is used as lower bound for \emph{partial string similarity} described in  the main paper.
\end{itemize} \par

\begin{algorithm*}[!ht]
\footnotesize
\SetAlgoLined
 potential\_matches = []\;
 \For{protagonist \textbf{in} protagonists}{
  ratio = regular\_string\_similarity(protagonist, named\_entity)\;
  \uIf{ratio == 100}{
   \textbf{return} match = protagonist\;
   }
  partial\_ratio = partial\_string\_similarity(protagonist, named\_entity)\;
  \uIf{partial\_ratio >= partial\_similarity\_precision}{
   potential\_matches\textbf{.add}(protagonist)\;
   }
  }
  potential\_matches = sorted(potential\_matches)  \# with respect to partial\_ratio \;
  match = \textbf{None} \;
  \uIf{len(potential\_matches) > 1}{
       match = potential\_matches[0]\;
       \uIf{prefix \textbf{is not None}}{
            \uIf{prefix == \emph{the}}{
            \textbf{return} match = \emph{the} + named\_entity}
            {
            title\_gender = get\_title\_gender(prefix) \# either female or male \;
             \For{potential\_match \textbf{in} potential\_matches}{
             \uIf{get\_name\_gender(potential\_match) == title\_gender}{
             \textbf{return} match = potential\_match}
             }
            }
       }
       \textbf{return} match
   }\uElseIf{len(potential\_matches) == 0}{
    original\_name = get\_name\_from\_diminutive(named\_entity)\;
    \uIf{original\_name \textbf{is not None}}{
    \textbf{return} match = protagonist from \emph{protagonists} that contains original\_name
    }
    \uElse{
    \textbf{return} "person"
   }
   }
   \textbf{return} potential\_match[0]
 \caption{Finding best match for the recognized named entity in the list of literary characters predefined for the analysed novel}
 \label{match_algorithm}
\end{algorithm*} 

The first step of the \textit{matching algorithm} is to check (using \emph{regular\_string\_similarity}) if the \emph{named\_entity} is identical to any of the literary characters from the \emph{protagonists'} list (lines 3-5 in Algorithm~\ref{match_algorithm}). If so, the algorithm ends and returns it as the best match. However, if it is not the case, the \emph{partial\_ratio} is computed for the \emph{named\_entity} and each literary character from the protagonists' list using \emph{partial\_string\_similarity}. If it is above the given threshold (\emph{partial\_similarity\_precision}), a given name from the protagonists' list is considered as a potential match (see lines 6-8 in Algorithm~\ref{match_algorithm}). The list of potential matches is sorted decreasingly concerning the computed \emph{partial\_ratio} (line 10). \par

At this stage, the only thing left to do is check whether the considered named entity is one of the exceptions that we are handling. First of all, we check whether the prefix (token preceding the recognized named entity) can give us any clue (see lines 14-21 in Algorithm~\ref{match_algorithm}). If the prefix is:
\begin{itemize}
    \item the article \emph{the} -- the whole family with the surname given in the named entity is considered;
    \item one of the personal titles -- prefix's gender is recognized, then the first literary character from \emph{potential\_matches} list that has the same gender as the personal title in the prefix is returned (see lines 17-21 in Algorithm~\ref{match_algorithm}). Here, we need to assume that the name higher in the list is more probable due to the higher similarity score. It may not always be the case, but some simplifications are necessary. 
\end{itemize} 
The last considered variant appears when not even a single literary character was qualified as a potential match. The reason for such a situation may be that the named entity includes not the basic form of the name of the literary character but the diminutive. In such a case, the additional search is performed in an external dictionary of diminutives containing more than 3300 different forms of names (see lines 23-26 in Algorithm~\ref{match_algorithm}). If the named entity is not found in the diminutive dictionary, a general tag \emph{person} is returned. It means that any of the predefined tags match the named entity.

\section{Detailed Statistics of NER models' and \textit{ProtagonistTagger}'s Performance} \label{appendix:statistics}

\begin{table*}
 \footnotesize
 \centering
 \setlength{\arrayrulewidth}{0.1mm}
\setlength{\tabcolsep}{3pt}
\renewcommand{\arraystretch}{1}
\begin{tabular}{ p{4cm}  p{1.5cm} | p{1.3cm}  p{1.3cm}  p{1.5cm}  p{1.3cm}}
 \multicolumn{2}{c|}{\textbf{Novel title / NER model}} & \textbf{Precision} & \textbf{Recall} & \textbf{F-measure} & \textbf{Support} \\
\hline\hline
\multicolumn{6}{c}{\textbf{\emph{Test\_large\_person}}} \\
\hline\hline
The Picture of Dorian Gray
    & standard &  0.69 &  0.41 &  0.51 &  90 \\
    & fine-tuned &  0.74 &  1    &  0.85 &  90 \\
\hline
Frankenstein
    & standard &  0.91 &  0.62 &  0.74 &  93 \\
    & fine-tuned  &  0.78 &  0.98 &  0.87 &  93 \\
\hline
Treasure Island
    & standard &  0.75 &  0.66 &  0.7  &  97 \\
    & fine-tuned &  0.78 &  1    &   0.87 &   97 \\
\hline
Emma
    & standard &  0.84 &  0.77 &  0.81 &  115 \\
    & fine-tuned &  0.85 &  1    &  0.92 &  115 \\
\hline
Jane Eyre
    & standard &  0.86 &  0.78 &  0.82 &  97 \\
    & fine-tuned &  0.74 &  0.95 &  0.83 &  97 \\
\hline
Wuthering Heights
    & standard &  0.95 &  0.87 &  0.91 &  108 \\
    & fine-tuned &  0.88 &  0.99 &  0.93 &  108 \\
\hline
Pride and Prejudice
    & standard &  0.85 &  0.87 &  0.86 &  124 \\
    & fine-tuned &  0.8  &  0.98 &  0.88 &  124 \\
\hline
Dracula
    & standard &  0.86 &  0.94 &  0.9  &  97 \\
    & fine-tuned &  0.72 &  0.99 &  0.83 &  97 \\
\hline
Anne of Green Gables
    & standard &  0.91 &  0.96 &  0.94 &  114 \\
    & fine-tuned &  0.85 &  0.99 &  0.92 &  114 \\
\hline
Adventures of Huckleberry Finn
    & standard &  0.71 &  0.99 &  0.83 &  86 \\
    & fine-tuned &  0.61 &  1    &  0.75 &  86 \\
\hline
\textbf{-- Overall results --}
    & \textbf{standard} &  \textbf{0.84} &  \textbf{0.8}  &  \textbf{0.82} &  \textbf{1021} \\
    & \textbf{fine-tuned} &  \textbf{0.77} &  \textbf{0.99} &  \textbf{0.87} &  \textbf{1021} \\
\hline\hline
\multicolumn{6}{c}{\textbf{\emph{Test\_small\_person}}} \\
\hline\hline
The Catcher in the Rye
    & standard &  0.68 &  0.68 &  0.68 &  74  \\
    & fine-tuned &  0.58 &  0.91 &  0.71 &  74 \\
\hline
The Great Gatsby
    & standard &  0.75 &  0.84 &  0.79 &  102 \\
    & fine-tuned &  0.66 &  0.98 &  0.79 &  102 \\
\hline
The Secret Garden
    & standard &  0.9 &  0.82 &  0.86 &  97  \\
    & fine-tuned &  0.83  &  0.95 &  0.88 &  97 \\
\hline
\textbf{-- Overall results --}
    & \textbf{standard}  &  \textbf{0.78} &  \textbf{0.79} &  \textbf{0.78} &  \textbf{273} \\
    & \textbf{fine-tuned} &  \textbf{0.69} &  \textbf{0.95} &  \textbf{0.8}  &  \textbf{273} \\
\end{tabular}
\caption{Metrics computed for the standard NER model and the fine-tuned NER model for annotations with general label \emph{person}. The \emph{support} is the number of occurrences (mentions) of class \emph{person}.}
\label{tab:app:metrics_fine_tuned_ner}
\end{table*}

Our prepared testing sets for NER models (pretrained and fine-tuned) comprises: \emph{Test\_large\_person} and \emph{Test\_small\_person}. These testing sets are manually annotated with general tag \emph{person} creating a gold standard for the NER model. The performance of \emph{protagonistTagger} is evaluated on \emph{Test\_large\_names} and \emph{Test\_small\_names}. Testing sets for \emph{protagonistTagger} include the same sentences as the corresponding sets used for testing the NER model. The only difference is that this time the sentences are manually annotated with full names of literary characters while creating the gold standard. \par

The performance of NER models (pretrained and fine-tuned) is evaluated on two testing sets: \emph{Test\_large\_person} and \emph{Test\_small\_person}. The overall performance of \emph{protagonistTagger} is evaluated on \emph{Test\_large\_names} and \emph{Test\_small\_names}.  \par

\begin{table*}
 \footnotesize
 \centering
 \setlength{\arrayrulewidth}{0.1mm}
\setlength{\tabcolsep}{3pt}
\renewcommand{\arraystretch}{1}
\begin{tabular}{ p{4cm}  p{1.5cm} p{1.5cm} p{1.5cm} }
 \textbf{Novel title} & \textbf{Precision} & \textbf{Recall} & \textbf{F-measure} \\
\hline
\multicolumn{4}{c}{\textbf{\emph{Test\_large\_names}}} \\
\hline
Pride and Prejudice             &        0.84 &     0.85 &        0.83 \\
 The Picture of Dorian Gray     &        0.96 &     0.97 &        0.96 \\
 Anne of Green Gables           &        0.94 &     0.96 &        0.95 \\
 Wuthering Heights              &        0.79 &     0.77 &        0.77 \\
 Jane Eyre                      &        0.8  &     0.75 &        0.76 \\
 Frankenstein                   &        0.91 &     0.88 &        0.89 \\
 Treasure Island                &        0.92 &     0.91 &        0.91 \\
 Adventures of Huckleberry Finn &        0.89 &     0.93 &        0.9  \\
 Emma                           &        0.93 &     0.86 &        0.88 \\
 Dracula                        &        0.9  &     0.89 &        0.89 \\
\textbf{-- Overall results --}    &    \textbf{0.88} &     \textbf{0.87} &        \textbf{0.87} \\
\hline
\multicolumn{4}{c}{\textbf{\emph{Test\_small\_names}}} \\
\hline
 The Catcher in the Rye  &        0.8  &     0.77 &        0.78 \\
 The Great Gatsby        &        0.88 &     0.9  &        0.89 \\
 The Secret Garden       &        0.8  &     0.79 &        0.79 \\
\textbf{-- Overall results --}    &    \textbf{0.83} &     \textbf{0.83} &       \textbf{ 0.83} \\
\end{tabular}
\caption{Performance of the \emph{protagonistTagger}.}
\label{tab:app:full_tags_large_set_overall_metrics}
\end{table*}

The metrics for the performance of the NER model are presented in~Table~\ref{tab:app:metrics_fine_tuned_ner} and the performance of the whole \emph{protagonistTagger} tool is presented in~Table~\ref{tab:app:full_tags_large_set_overall_metrics}. All the metrics are given for each novel individually and for each testing set in general.

\section{Detailed Statistics for Problems Handled by the \textit{Matching Algorithm}}\label{stats_for_matching_alg_problems}
The most problematic cases in the matching process are diminutives and nicknames. The problem accompanies the character \emph{Elizabeth Bennet}, who is sometimes called \emph{Lizzy} by her family. Statistics presented in~Table~\ref{tab:Elizabeth} shows that in the case of \emph{Pride and Prejudice} by Jane Austen this problem is quite common. In order to discover a base form of a diminutive detected in a text, we use the external list containing the most common variants of names.  \par

\begin{table}
  \centering
  \footnotesize
    \begin{tabular}{ p{2.5cm}  p{2.5cm} }
        Named entity & \# of appearances \\ 
        \hline
         Elizabeth & 635 \\
         Lizzy & 96 \\ 
         Miss Bennet & 72 \\
         Miss Elizabeth & 12 \\ 
         Elizabeth Bennet & 8 \\ 
    \end{tabular}
    \caption{Appearances of the references to \emph{Elizabeth Bennet} in \emph{Pride and Prejudice} in different forms.}
    \label{tab:Elizabeth}
\end{table}

Another case that needs special consideration is a named entity consisting only of a surname. We consider this situation on the example of \emph{Bennet} named entity. As it is discussed in the paper, we want to distinguish between \emph{Bennet} meaning the whole family and \emph{Bennet} being the surname of a single character. We can do it by analyzing the word preceding the detected named entity. \emph{Bennet} preceded with a personal title such as Mr., Mrs., Ms. or Miss, should be identified as a single person, whose surname is \emph{Bennet}. In all other cases, \emph{Bennet} is treated as the whole family and not a single person identified in a text. The statistics presented in~Table~\ref{tab:Bennet} illustrates the scale of the problem in this specific novel, which has 121,533 words. The entity \emph{Bennet} appears 323 times, out of which 314 cases can be analyzed more precisely thanks to the preceding personal title.

\begin{table}
    \centering
    \footnotesize
    \begin{tabular}{ p{2.5cm}  p{2.5cm} }
        Named entity & \# of appearances \\
        \hline
         Bennet & 323 \\ 
         Mrs. Bennet & 153 \\ 
         Mr. Bennet & 89 \\ 
         Miss Bennet & 72 \\
    \end{tabular}
    \caption{Appearances of the named entity \emph{Bennet} in \emph{Pride and Prejudice} with different personal titles.}
    \label{tab:Bennet}
\end{table}

\section{\textit{ProtagonistTagger}'s Performance vs Number of Literary Characters}
\label{appendix:results_analysis}
\begin{figure*}
    \centering
    \includegraphics[width=0.9\textwidth]{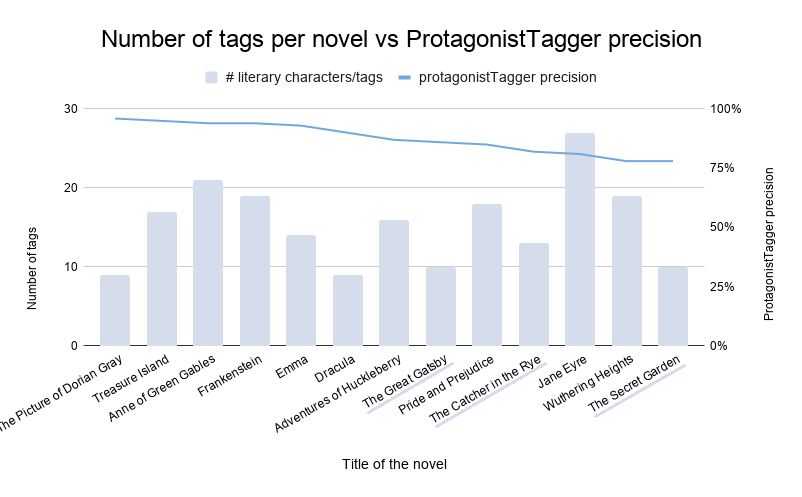}
    \caption{The left vertical axis describes the number of literary characters (tags used by the \emph{protagonistTagger}) in each novel, whereas the right vertical axis describes the precision of the \emph{protagonistTagger} (given in percents) for each novel. Novels used in \emph{Test\_small\_names} are marked with grey underlining.}
    \label{fig:names_amouts_graph}
\end{figure*}

One of the factors that intuitively should influence the performance of the \emph{protagonistTagger} is the number of literary characters in a novel that is analysed. The number of protagonists in a novel determines the number of tags that are used by the \emph{protagonistTagger}. The more tags the tool has to choose from, the more difficult is the task of matching them correctly to each recognized named entity. The relation between the precision of the \emph{protagonistTagger} and the number of tags per each novel is presented in~Figure~\ref{fig:names_amouts_graph}. It can not be said unambiguously that these two values are in inverse proportion in the analysed testing sets. The novels for which the tool achieved both the lowest and the highest precision -- \emph{The Picture of Dorian Gray} and \emph{The Secret Garden} -- have a relatively small number of literary characters. \par 

\begin{figure*}
    \centering
    \includegraphics[width=0.9\textwidth]{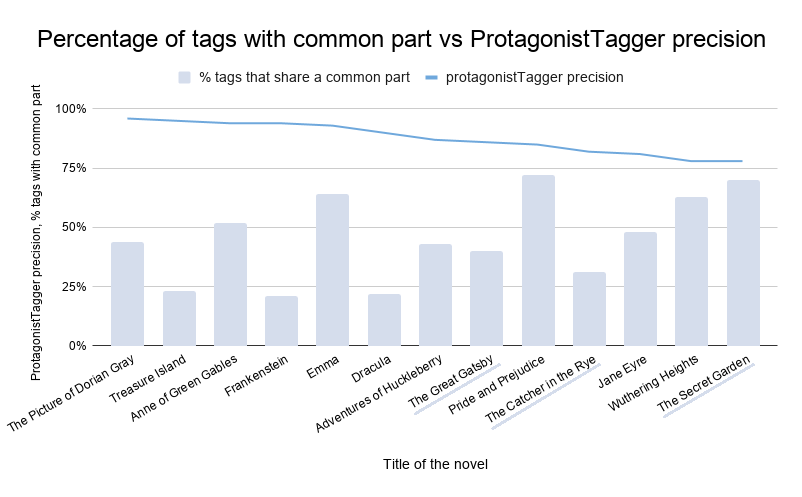}
    \caption{The right vertical axis describes the percentage of tags sharing a common part 
    (grey bars), as well as the precision of the \emph{protagonistTagger} (given in percents) for each novel in the testing sets. Novels used in \emph{Test\_small\_names} are marked with grey underlining.}
    \label{fig:names_common_parts_graph}
\end{figure*}

Another factor that was suspected to negatively influence the performance of the \emph{protagonistTagger} is the number of tags sharing a common part. In the case of \emph{Bennet} family in \emph{Pride and Prejudice} by Jane Austen protagonists with the same surname are problematic even for human annotators. Sometimes a personal title preceding the named entity can be helpful. However, matching correctly tags that share the same surname or even name may be nontrivial. For that reason, we created statistics of tags that share some common part. These statistics, presented in~Table~\ref{tab:names_amounts}, are given for each novel included in both testing sets. Additionally, they are presented in~Figure~\ref{fig:names_common_parts_graph} along with the performance of the \emph{protagonistTagger}.

\begin{table*}
 \footnotesize
 \centering
 \setlength{\arrayrulewidth}{0.1mm}
\setlength{\tabcolsep}{3pt}
\renewcommand{\arraystretch}{1}
    \begin{tabular}{  p{5cm} | p{2.5cm} | p{2.5cm} | p{2.5cm}  }
    Title of the novel & \# literary characters/tags & \# tags that share a common part & \% tags that share a common part \\ [0.5ex]
    \hline
     Pride and Prejudice            & 18 & 13 & 72\% \\
     The Picture of Dorian Gray     &  9 &  4 & 44\% \\
     Anne of Green Gables           & 21 & 11 & 52\% \\
     Wuthering Heights              & 19 & 12 & 63\% \\
     Jane Eyre                      & 27 & 13 & 48\% \\
     Frankenstein                   & 19 &  4 & 21\% \\
     Treasure Island                & 17 &  4 & 23\% \\
     Adventures of Huckleberry Finn & 16 &  7 & 43\% \\
     Emma                           & 14 &  9 & 64\% \\
     Dracula                        &  9 &  2 & 22\% \\
     \hline
     The Catcher in the Rye         & 13 &  4 & 31\% \\
     The Great Gatsby               & 10 &  4 & 40\% \\
     The Secret Garden              & 10 &  7 & 70\% \\
    \end{tabular}
    \caption{The number of literary characters (tags used by \emph{protagonistTagger}) appearing in each novel and the number of tags that share a common part. A common part can be the same name or surname. The same personal title in two tags is not considered a common part.}
    \label{tab:names_amounts}
\end{table*}

\begin{table*}
 \footnotesize
 \centering
 \setlength{\arrayrulewidth}{0.1mm}
\setlength{\tabcolsep}{3pt}
\renewcommand{\arraystretch}{1}
    \begin{tabular}{  p{4cm} | p{11cm} }
    Parameter or component & Value and/or description \\ [0.5ex]
    \hline
    pretrained Spacy model & en\_core\_web\_sm (\url{https://spacy.io/models/en\#en\_core\_web\_sm})\\
        word features & Bloom word embeddings with sub word features \\
    pretrained model architecture & deep convolution neural network with residual connections \\
    fine-tuning iterations & 100 \\
    dropout rate & 0.5 \\
    batch normalization & minibatch (size - a series of compounding values starting at 4.0, stopping at 32.0, with compound equal to 1.001) \\
    optimizer & Adam (learning rate=0.001; beta1=0.9; beta2=0.999; eps=1e-08; L2=1e-6; grad\_clip=1.0; use\_averages=True; L2\_is\_weight\_decay=True) \\
    \end{tabular}
    \caption{Details about the pretrained NER model and parameters used for the fine-tuning procedure.}
    \label{tab:ner_params}
\end{table*}

However, again no obvious relation between these two values is visible. It may be caused by the fact that common parts in tags may not be related to the main protagonists (the ones that appear most often in the novel and the testing sets). Therefore, the testing sets are not representative enough in this case. For example, in the case of \emph{Anne of Green Gables} that has relatively many literary characters, half of which share the same name or surname, the tool's performance is very high. Nonetheless, in the case of \emph{Wuthering Heights}, with a similar number of protagonists, half of which again share a common name or surname, performance is much lower. It is caused by the fact that in \emph{Wuthering Heights} the tags that share common parts correspond to the main protagonists. Whereas, in the case of \emph{Anne of Green Gables} such common elements appear rather in tags corresponding to tangential characters. \par

In general, it can be concluded that the performance of the \emph{protagonistTagger} depends on many factors, not only the number of tags and the percentage of tags with the common part in a novel. These two factors, in some cases, can negatively influence the performance of the tool. However, this impact is not certain in the case of all novels.

\section{Details for Reproducing the Experiments}
\label{appendix:models_hyperparams}

The documented code along with the attached manual that is provided with this paper is the best starting point for reproducing experiments. The created scripts allow to repeat all the actions and run all the tests described in the paper. Annotated corpus of thirteen novels is provided as a part of the code. Nevertheless, it is possible to expand it with new annotated texts using the provided scrips (being a part of \textit{protagonistTagger}). Table~\ref{tab:ner_params} contains the summary of the parameters and techniques applied to fine-tune the pretrained NER model~\footnote{\footnotesize{\url{https://spacy.io/models/en\#en\_core\_web\_sm}}}. \par

The pretrained NER model uses embeddings with subwords features, convolutional layers with residual connections, layer normalization, and maxout non-linearity (its output is the max of a set of inputs) \cite{maxout}. The training data is shuffled and batched. For each batch, the model is updated with the training sentences from a batch. The dropout is applied as a regularisation technique to make it a little bit harder for the model to memorize data and reduce overfitting.

In order to successfully use scripts provided as a part of \emph{protagonistTagger} tool the following requirements need to be fulfilled:
\begin{itemize}
    \item Python 3.6
    \item PyYAML 5.3
    \item gensim 3.8
    \item numpy 1.18.2
    \item pytorch.transformers 1.2
    \item scikit-learn 0.22
    \item scipy 1.4.1
    \item spacy 2.2.4
\end{itemize}

Additionally the following external packages are used:
\begin{itemize}
    \item fuzzywuzzy 0.18~\footnote{\footnotesize{\url{https://pypi.org/project/fuzzywuzzy/}}}
    \item gender-guesser 0.4~\footnote{\footnotesize{\url{https://pypi.org/project/gender-guesser/}}}
    \item nickname-and-diminutive-names-lookup~\footnote{\footnotesize{\url{https://github.com/carltonnorthern/nickname-and-diminutive-names-lookup}}}
\end{itemize}

\end{document}